# A navigation filter for fusing DTM/correspondence updates.

Oleg Kupervasser, Vladimir Voronov

Transas group of companies
Moscow, Russia
Email: olegkup@yahoo.com

August 2011

Abstract

An algorithm for pose and motion estimation using corresponding features in images and a digital terrain map is proposed. Using a Digital Terrain (or Digital Elevation) Map (DTM/DEM) as a global reference enables recovering the absolute position and orientation of the camera. In order to do this, the DTM is used to formulate a constraint between corresponding features in two consecutive frames. The utilization of data is shown to improve the robustness and accuracy of the inertial navigation algorithm. Extended Kalman filter was used to combine results of inertial navigation algorithm and proposed vision-based navigation algorithm. The feasibility of this algorithms is established through numerical simulations.

## 0.1 Introduction.

Vision-based algorithms has been a major research issue during the past decades. Two common approaches for the navigation problem are: landmarks and ego-motion integration. In the landmarks approach several features are located on the image-plane and matched to their known 3D location. Using the 2D and 3D data the camera's pose can be derived. Few examples for such algorithms are [2], [3]. Once the landmarks were found, the pose derivation is simple and can achieve quite accurate estimates. The main difficulty is the detection of the features and their correct matching to the landmarks set.

In ego-motion integration approach the motion of the camera with respect to itself is estimated. The ego-motion can be derived from the optical-flow field, or from instruments such as accelerometers and gyroscopes. Once the ego-motion was obtained, one can integrate this motion to derive the camera's path. One of the factors that make this approach attractive is that no specific features need to be detected, unlike the previous approach. Several ego-motion estimation algorithms can be found in [4], [5], [6], [7]. The weakness of ego-motion integration comes from the fact that small errors are accumulated during the integration process. Hence, the estimated camera's path is drifted and the pose estimation accuracy decrease along time. If such approach is used it would be desirable to reduce the drift by activating, once in a while, an additional algorithm that estimates the pose directly. In [8], such navigation-system is being suggested. In that work, like in this work, the drift is being corrected using a Digital Terrain Map (DTM). The DTM is a discrete representation of the observed ground's topography. It contains the altitude over the sea level of the terrain for each geographical location. In [8] a patch from the ground was reconstructed using 'structure-from-motion' (SFM) algorithm and was matched to the DTM in order to derive the camera's pose. Using SFM algorithm which does not make any use of the information obtained from the DTM but rather bases its estimate on the flow-field alone, positions their technique under the same critique that applies for SFM algorithms [1].

The algorithm presented in this work does not require an intermediate explicit reconstruction of the 3D world. By combining the DTM information directly with the images information it is claimed that the algorithm is well-conditioned and generates accurate estimates for reasonable scenarios and error sources. In the present work this claim is explored by performing an error analysis on the algorithm outlined above. By assuming appropriate characterization of these error sources, a closed form expression for the uncertainty of the pose and motion of the camera is first developed and then the influence of different factors is studied using extensive numerical simulations.

## 0.2 Problem Definition and Notations.

The problem can be briefly described as follows: At any given time instance $t$, a coordinates system $C(t)$ is fixed to a camera in such a way that the $Z$-axis coincides with the optical-axis and the origin coincides with the camera's projection center. At that time instance the camera is located at some geographical location $p(t)$ and has a given orientation $R(t)$ with respect to a global coordinates system $W$ ($p(t)$ is a 3D vector, $R(t)$ is an orthonormal rotation matrix). $p(t)$ and $R(t)$ define the transformation from the camera's frame $C(t)$ to the world's frame $W$, where if $^Cv$ and $^Wv$ are vectors in $C(t)$ and $W$ respectively, then $^Wv = R(t)^Cv + p(t)$.

Consider now two sequential time instances $t_1$ and $t_2$: the transformation from $C(t_1)$ to $C(t_2)$ is given by the translation vector $\Delta p(t_1, t_2)$ and the rotation matrix $\Delta R(t_1, t_2)$, such that $^{C(t_2)}v = \Delta R(t_1, t_2)\,^{C(t_1)}v + \Delta p(t_1, t_2)$. A rough estimate of the camera's pose at $t_1$ and of the ego-motion between the two time instances - $p_E(t_1)$, $R_E(t_1)$, $\Delta p_E(t_1, t_2)$ and $\Delta R_E(t_1, t_2)$ - are supplied (the subscript letter "$E$" denotes that this is an estimated quantity).

Also supplied is the optical-flow field: $\{u_i(t_k)\}$ (i=1...n, k=1,2). For the $i$'th feature, $u_i(t_1) \in \mathbb{R}^2$ and $u_i(t_2) \in \mathbb{R}^2$ represent its locations at the first and second frame respectively.

Using the above notations, the objective of the proposed algorithm is to estimate the true camera's pose and ego-motion: $p(t_1)$, $R(t_1)$, $\Delta p(t_1, t_2)$ and $\Delta R(t_1, t_2)$, using the optical-flow field $\{u_i(t_k)\}$, the DTM and the initial-guess: $p_E(t_1)$, $R_E(t_1)$, $\Delta p_E(t_1, t_2)$ and $\Delta R_E(t_1, t_2)$.

## 0.3 The Navigation Algorithm

The following section describes a navigation algorithm which estimate the above mentioned parameters. The pose and ego-motion of the camera are derived using a DTM and the optical-flow field of two consecutive frames. Unlike the landmarks approach no specific features should be detected and matched. Only the correspondence between the two consecutive images should be found in order to derive the optical-flow field. As was mentioned in the previous section, a rough estimate of the required parameters is supplied as an input. Nevertheless, since the algorithm only use this input as an initial guess and re-calculate the pose and ego-motion directly, no integration of previous errors will take place and accuracy will be preserved.

The new approach is founded on the following observation. Since the DTM supplies information about the structure of the observed terrain, depth of observed features is being dictated by the camera's pose. Hence, given the pose and ego-motion of the camera, the optical-flow field can be uniquely determined. The objective of the algorithm will be finding the pose and ego-motion which lead to an optical-flow field as close as possible to the given flow field.

A single vector from the optical-flow field will be used to define a constraint for the camera's pose and ego-motion. Let $^WG \in \mathbb{R}^3$ be a location of a ground feature point in the 3D world. At two different time instances $t_1$ and $t_2$, this feature point is projected on the image-plane of the camera to the points $u(t_1)$ and $u(t_2)$. Assuming a pinhole model for the camera, then $u(t_1), u(t_2) \in \mathbb{R}^2$. Let $^Cq(t_1)$ and $^Cq(t_2)$ be the homogeneous representations of these locations. As standard, one can think of these vectors as the vectors from the optical-center of the camera to the projection point on the image plane. Using an initial-guess of the pose of the camera at $t_1$, the line passing through $p_E(t_1)$ and $^Cq(t_1)$ can be intersected with the DTM. Any ray-tracing style algorithm can be used for this purpose. The location of this intersection is denoted as $^WG_E$. The subscript letter "$E$" highlights the fact that this ground-point is the estimated location for the feature point, that in general will be different from the true ground-feature location $^WG$. The difference between the true and estimated locations is due to two main sources: the error in the initial guess for the pose and the errors in the determination of $^WG_E$ caused by DTM discretization and intrinsic errors. For a reasonable initial-guess and DTM-related errors, the two points $^WG_E$ and $^WG$ will be close enough so as to allow the linearization of the DTM around $^WG_E$. Denoting by $N$ the normal of the plane tangent to the DTM at the point $^WG_E$, one can write:

$$N^T(^WG - {}^WG_E) \approx 0 \tag{1}$$

The true ground feature $^W G$ can be described using true pose parameters:

$$^W G = R(t_1) \cdot {}^C q(t_1) \cdot \lambda + p(t_1) \qquad (2)$$

Here, $\lambda$ denotes the depth of the feature point (i.e. the distance of the point to the image plane projected on the optical-axis). Replacing (2) in (1):

$$N^T (\lambda \cdot R(t_1) \cdot {}^C q(t_1) + p(t_1) - {}^W G_E) = 0 \qquad (3)$$

From this expression, the depth of the true feature can be computed using the estimated feature location:

$$\lambda = \frac{N^{T\,W} G_E - N^T p(t_1)}{N^T R(t_1) {}^C q(t_1)} \qquad (4)$$

By plugging (4) back into (2) one gets:

$$^W G = R(t_1) {}^C q(t_1) \cdot \left( \frac{N^{T\,W} G_E - N^T p(t_1)}{N^T R(t_1) {}^C q(t_1)} \right) + p(t_1) \qquad (5)$$

In order to simplify notations, $R(t_i)$ will be replaced by $R_i$ and likewise for $p(t_i)$ and $q(t_i)$ $i = 1, 2$. $\Delta R(t_1, t_2)$ and $\Delta p(t_1, t_2)$ will be replaced by $R_{12}$ and $p_{12}$ respectively. The superscript describing the coordinate frame in which the vector is given will also be omitted, except for the cases were special attention needs to be drawn to the frames. Normally, $p_{12}$ and $q$'s are in camera's frame while the rest of the vectors are given in the world's frame. Using the simplified notations, (5) can be rewritten as:

$$G = \frac{R_1 q_1 N^T}{N^T R_1 q_1} G_E - \frac{R_1 q_1 N^T}{N^T R_1 q_1} p_1 + p_1 \qquad (6)$$

In order to obtain simpler expressions, define the following projection operator:

$$\mathcal{P}(u, s) \doteq \left( I - \frac{u s^T}{s^T u} \right) \qquad (7)$$

This operator projects a vector onto the subspace normal to $s$, along the direction of $u$. As an illustration, it is easy to verify that $s^T \cdot \mathcal{P}(u, s) v \equiv 0$ and $\mathcal{P}(u, s) u \equiv 0$. By adding and subtracting $G_E$ to (6), and after reordering:

$$G = G_E + \left[ I - \frac{R_1 q_1 N^T}{N^T R_1 q_1} \right] p_1 - \left[ I - \frac{R_1 q_1 N^T}{N^T R_1 q_1} \right] G_E \qquad (8)$$

Using the projection operator, (8) becomes:

$$G = G_E + \mathcal{P}(R_1 q_1, N)(p_1 - G_E) \qquad (9)$$

The above expression has a clear geometric interpretation (see Fig.1). The vector from $G_E$ to $p_1$ is being projected onto the tangent plane. The projection is along the direction $R_1 q_1$, which is the direction of the ray from the camera's optical-center ($p_1$), passing through the image feature.

Our next step will be transferring $G$ from the global coordinates frame- $W$ into the first camera's frame $C_1$ and then to the second camera's frame $C_2$. Since $p_1$ and $R_1$ describe the transformation from $C_1$ into $W$, we will use the inverse transformation:

$$^{C_2} G = p_{12} + R_{12} \left( R_1^T (G - p_1) \right) \qquad (10)$$

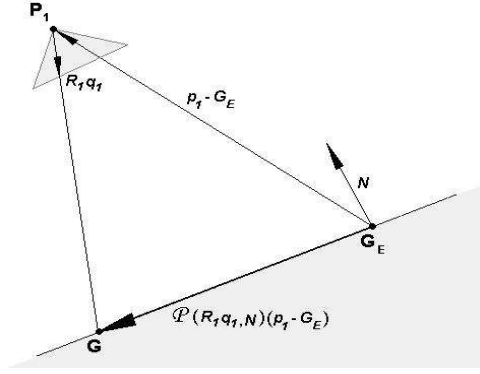

Figure 1: Geometrical description of expression (9) using the projection operator (7)

Assigning (9) into (10) gives:

$$^{C_2}G = p_{12} + R_{12}\mathcal{L}\left(G_E - p_1\right) \qquad (11)$$

$\mathcal{L}$ in the above expression represents:

$$\mathcal{L} = \frac{q_1 N^T}{N^T R_1 q_1} \qquad (12)$$

One can think of $\mathcal{L}$ as an operator with inverse characteristic to $\mathcal{P}$: it projects vectors on the ray continuing $R_1 q_1$ along the plane orthogonal to $N$.

$q_2$ is the projection of the true ground-feature $G$. Thus, the vectors $q_2$ and $^{C_2}G$ should coincide. This observation can be expressed mathematically by projecting $^{C_2}G$ on the ray continuation of $q_2$:

$$^{C_2}G = \frac{q_2}{|q_2|} \cdot \left(\frac{q_2^T}{|q_2|} \cdot {}^{C_2}G\right) \qquad (13)$$

In expression (13), $q_2^T/|q_2| \cdot {}^{C_2}G$ is the magnitude of $^{C_2}G$'s projection on $q_2$. By reorganizing (13) and using the projection operator, we obtain:

$$\left[\mathrm{I} - \frac{q_2 \cdot q_2^T}{q_2^T \cdot q_2}\right] \cdot {}^{C_2}G = \mathcal{P}(q_2, q_2) \cdot {}^{C_2}G = 0 \qquad (14)$$

$^{C_2}G$ is being projected on the orthogonal complement of $q_2$. Since $^{C_2}G$ and $q_2$ should coincide, this projection should yield the zero-vector. Plugging (11) into (14) yields our final constraint:

$$\mathcal{P}(q_2, q_2)\left[p_{12} + R_{12}\mathcal{L}\left(G_E - p_1\right)\right] = 0 \qquad (15)$$

This constraint involves the position, orientation and the ego-motion defining the two frames of the camera. Although it involves 3D vectors, it is clear that its rank can not exceed two due to the usage of $\mathcal{P}$ which projects $\mathbb{R}^3$ on a two-dimensional subspace.

Such constraint can be established for each vector in the optical-flow field, until a non-singular system is obtained. Since twelve parameters need to be estimated (six for pose and six for the ego-motion), at least six optical-flow vectors are required for the system solution. But it is correct conclusion for nonlinear problem. If we use Gauss-Newton iterations method and so make linearization of our problem near approximate solution. The found matrix will be always singular for six points (with zero determinant)as numerical simulations demonstrate. So it is necessary to use at least seven points to obtain nonsingular linear approximation. Usually, more vectors will be used in order to define an over-determined system, which will lead to more robust solution. The reader attention is drawn to the fact that a non-linear constraint was obtained. Thus, an iterative scheme will be used in order to solve this system. A robust algorithm which uses Gauss-Newton iterations and M-estimator is described in [9].We begin to use Levenberg-Marquardt method if Gauss-Newton method after several iterations stopped to converge. This two algorithms are realized in lsqnonlin() Matlab function. The applicability, accuracy and robustness of the algorithm was verified though simulations and lab-experiments.

It is more convenient to use more robust for iterations equivalent to (15) equation:

$$\mathcal{P}(q_2, q_2)\left[p_{12} + R_{12}\mathcal{L}_i\left(G_{E_i} - p_1\right)\right]/|^{C_2}G| = 0 \qquad (16)$$

Using of this normalized form of equations avoids to get incorrect trivial solution when two positions are in a single point on the ground.

## 0.4 Vision-based navigation algorithm corrections for inertial navigation by help of Kalman filter.

Vision-based navigation algorithms has been a major research issue during the past decades. Algorithm used in this paper is based on foundations of multiple-view geometry and a land map. By help of this method we get position and orientation of a observer camera. On the other hand we obtain the same data from inertial navigation methods. To adjust these two results Kalman filter is used. We employ in this paper extended Kalman filter for nonlinear equations [12].

For inertial navigation computations was used Inertial Navigation System Toolbox for Matlab [13].

Input of Kalman filter consists of two part. The first one is variables $X$ for equations of motion. In our case it is inertial navigation equations. Vector $X$ consists of fifteen components: $[\delta x\ \delta y\ \delta z\ \delta V_x\ \delta V_y\ \delta V_z\ \delta\phi\ \delta\theta\ \delta\psi\ a_x\ a_y\ a_z\ b_x\ b_y\ b_z]$. Coordinates $\delta x\ \delta y\ \delta z$ are defined by difference between real position of the camera and position gotten from inertial navigation calculus.Variables $\delta V_x \delta V_y \delta V_z$ are defined by difference between real velocity of the camera and velocity gotten from inertial navigation calculus. Variable $\delta\phi\ \delta\theta\ \delta\psi$ are defined as Euler angles of matrix $D_r * D_c^T$ where $D_r$ is matrix defined by real Euler angles of camera with respect to Local Level Frame (L-Frame) and $D_c$ is matrix defined by Euler angles of camera with respect to Local Level Frame (L-Frame) gotten by inertial navigation computation. It is necessary to pay attention that found Euler angles $\delta\phi\delta\theta\delta\psi$ ARE NOT equivalent to difference between real Euler angles and Euler angles gotten from inertial navigation calculus. For small values of $\delta\phi\ \delta\theta\ \delta\psi$ perturbations to these angles can be added linearly and so these angles can be used in Kalman filter for small errors. Such choose of angles is made because formulas describing their evolution are much simpler than formulas describing evolution of Euler angles differences. Variables $a_x\ a_y\ a_z$ are defined by vector of Accel bias in inertial navigation measurements.Variables $b_x b_y b_z$ are defined by vector of Gyro bias in inertial navigation measurements.

The second input of Kalman filter is $Z$-result of measurements by vision-based navigation algorithms.Vector $Z$ consists of six components $[\delta x_m \delta y_m \delta z_m \delta\phi_m \delta\theta_m \delta\psi_m]$Coordinates $\delta x_m \delta y_m \delta z_m$ are difference between camera position measured by vision-based navigation algorithm and position gotten from inertial navigation calculus.Variable $\delta\theta_m \delta\psi_m$ are defined as Euler angles of matrix $D_m * D_c^T$ where $D_m$ is matrix defined by Euler angles of camera with respect to Local Level Frame (L-Frame) measured by vision-based navigation algorithm and $D_c$ is matrix defined by Euler angles of camera with respect to Local Level Frame (L-Frame) gotten by inertial navigation computation. Let variable $k$ to be number of step for time discretization used in Kalman filter.

We assume that errors between values gotten by inertial navigation computation and real values are linearly depend on noise. Corespondent process noise covariance matrix is denoted by $Q_k$. Diagonal elements of $Q_k$ correspondent to velocity are defined by Accel noise and proportional to $dt^2$: $Q_V \sim dt^2$, where $dt$ is time interval between $t_k$ and $t_{k-1}$: $dt = t_k - t_{k-1}$. Diagonal elements of $Q_k$ correspondent to Euler angles are defined by Gyro noise and proportional to $dt$: $Q_A \sim dt$.

We assume that errors between values gotten by vision-based navigation algorithm and real values are linearly depend on noise. Corespondent measurement noise covariance matrix is denoted by $R_k$. Error analysis giving this matrix is described in [14].

Kalman filter equations describe evolution of *a posteriori* state estimation $X_k$ described above and *aposteriori* error covariation covariance matrix $P_k$ for variables $X_k$.

To write Kalman filter equations we must define two 15x15 matrices yet: $H_k$ and $A_k$. Matrix $H_k$ is measurement Jacobian describing connection between predicted measurement $H_k * X_k$ and actual measurement $Z_k$ defined above. Diagonal elements $H_k(1,1)$, $H_k(2,2)$, $H_k(3,3)$ describing coordinate and elements $H_k(4,7)$, $H_k(5,8)$, $H_k(6,9)$ describing angles are equal to one. The rest of the elements are equal to zero.

$A_k$ is Jacobian matrix describing evolution of vector $X_k$. The exact expression for this matrix is very difficult so we use approximate formula for $A_k$ neglecting by Coriolis effects, Earth rotation and so on. Let $\phi$ $\theta$ $\psi$ be the Euler angles in L-Frame, $dV$ is deltaV vector gotten from inertial navigation measurements, $f_{vec}$ is acceleration vector in L-frame, $DCM_{\text{b-to-l}}$ is direction cosine matrix (from body-frame to L-frame).

The formulas defining $A_k$ are follow:

$$\Psi_{DCM} = \begin{pmatrix} \cos(\psi) & \sin(\psi) & 0 \\ -\sin(\psi) & \cos(\psi) & 0 \\ 0 & 0 & 1 \end{pmatrix} \tag{17}$$

$$\Theta_{DCM} = \begin{pmatrix} \cos(\theta) & 0 & -\sin(\theta) \\ 0 & 1 & 0 \\ \sin(\theta) & 0 & \cos(\theta) \end{pmatrix} \tag{18}$$

$$\Phi_{DCM} = \begin{pmatrix} 1 & 0 & 0 \\ 0 & \cos(\phi) & \sin(\phi) \\ 0 & -\sin(\phi) & \cos(\phi) \end{pmatrix} \tag{19}$$

$$DCM_{\text{b-to-l}} = \Phi_{DCM} \Theta_{DCM} \Psi_{DCM} \tag{20}$$

$$f_{vec} = DCM_{\text{b-to-l}} \frac{dV}{dt} \tag{21}$$

$$Phi(1:3, 4:6) = \begin{pmatrix} 1 & 0 & 0 \\ 0 & 1 & 0 \\ 0 & 0 & 1 \end{pmatrix} \tag{22}$$

$$Phi(4:6, 7:9) = \begin{pmatrix} 0 & -f_{vec}(3) & f_{vec}(2) \\ f_{vec}(3) & 0 & -f_{vec}(1) \\ -f_{vec}(2) & f_{vec}(1) & 0 \end{pmatrix} \quad (23)$$

$$Phi(7:9, 10:12) = -DCM_{\text{b-to-l}} \quad (24)$$

$$Phi(4:6, 13:15) = -DCM_{\text{b-to-l}} \quad (25)$$

The rest of elements for matrix Phi are equal to zero.

$$A_k = I + Phi\, dt \quad (26)$$

Kalman filter time update equations are follow:

$$X_k^- = [0\,0\,0\,0\,0\,0\,0\,0\,0\, a_{xk-1}\, a_{yk-1}\, a_{zk-1}\, b_{xk-1}\, b_{yk-1}\, b_{zk-1}] \quad (27)$$

$$P_k^- = A_k P_{k-1} A_k^T + Q_{k-1} \quad (28)$$

Kalman filter update equations project the state and covariance estimates from the previous time step $k-1$ to the current time step $k$.

Kalman filter measurement update equations are follow:

$$K_k = P_k^- H_k^T (H_k P_k^- H_k^T + R_k)^{-1} \quad (29)$$

$$X_k = X_k^- + K_k(Z_k - H_k X_k^-) \quad (30)$$

$$P_k = (I - K_k H_k) P_k^- (I - K_k H_k)^T + K_k R_k K_k^T \quad (31)$$

Kalman filter measurement update equations correct the state and covariance estimates with measurement $Z_k$.

The found vector $X_k$ is used to update coordinates, velocities, Euler angles, Accel and Gyro biases for inertial navigation calculations on the next step.

Numerical simulations were realized to examine effectiveness of Kalman filter to combine these two navigation algorithms. On Fig. 2 we can see that corrected path for coordinate error much smaller than inertial navigation coordinate error without Kalman filter. Improved results by help Kalman filter are gotten also for velocity in spite of the fact that this velocity was not measured by help vision-based navigation algorithm Fig. 3.

## 0.5 Conclusions

An algorithm for pose and motion estimation using corresponding features in images and a DTM was presented with using Kalman filter. The DTM served as a global reference and its data was used for recovering the absolute position and orientation of the camera. In numerical simulations position and velocity estimates were found to be sufficiently accurate in order to bound the accumulated errors and to prevent trajectory drifts.

## Acknowledgment

We would like to thank Ronen Lerner, Ehud Rivlin and Hector Rotstein for very useful consultations.

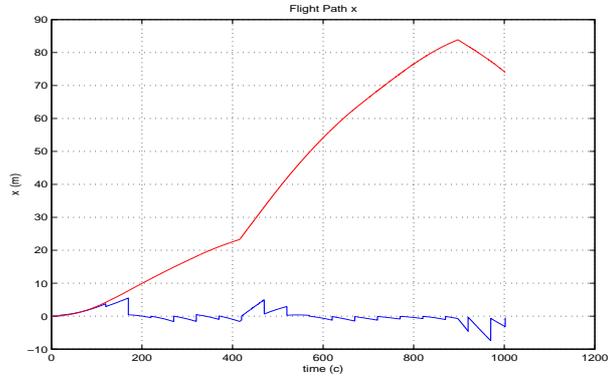

(a)

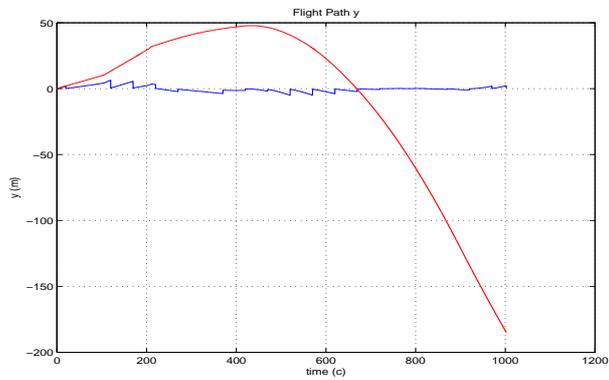

(b)

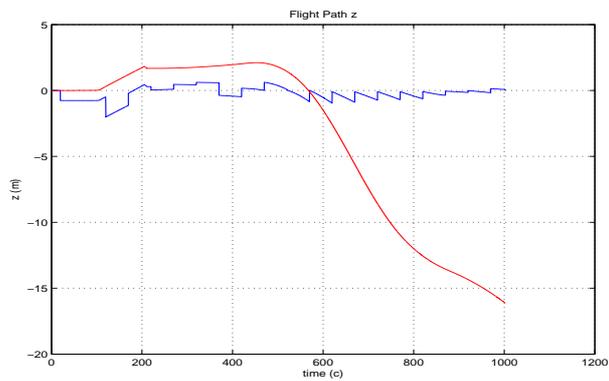

(c)

Figure 2: Position errors ((a) for x coordinate (b) for y coordinate (c) for z coordinate) of the drift path are marked with a red line, and errors of the corrected path are marked with a blue line.

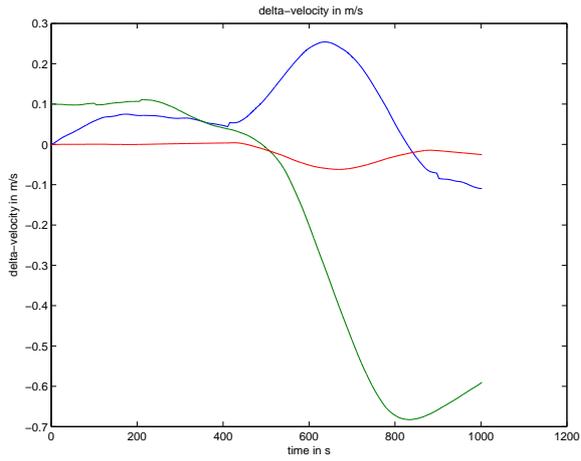

(a)

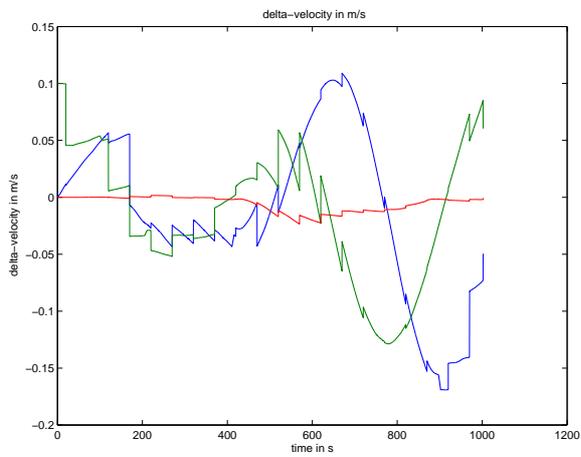

(b)

Figure 3: (a) Velocity errors of the drift path (x y z components), and (b) Velocity errors of the corrected path (x y z components).

# Навигационный фильтр Калмана для коррекции ошибок инерциальной навигации с помощью Цифровой Карты Ландшафта и соответствующих точек на кадрах фильма.

Купервассер Олег Юрьевич, Воронов Владимир Владимирович

группа компаний Транзас
Москва, Россия
Email: olegkup@yahoo.com

Август 2011

## Аннотация

Предложен алгоритм для нахождения позиции, ориентации и оценки движения, использующий соответствующие точки в изображениях и цифровую карту ландшафта. Использование Цифровой Карты Ландшафта (ЦКЛ) как глобальной справочной информации позволяет восстановление абсолютной позиции и ориентации камеры. Чтобы сделать это, ЦКЛ используется, чтобы сформулировать дополнительные ограничения между соответствующими точками в двух последовательных кадрах. Использование этих данных позволяет улучшить надежность и точность инерциального навигационного алгоритма. Расширенный фильтр Калмана использовался, чтобы объединить результаты инерциального навигационного алгоритма и навигационного алгоритма, основанного на компьютерном зрении. Выполнимость этого алгоритма продемонстрирована путем численного моделирования.

## 0.1 Введение.

Основанные на системе технического зрения алгоритмы были главной исследовательской проблемой в течение прошлых десятилетий. Два единых подхода для навигационной проблемы существуют: наземные ориентиры и интеграция собственного движения. В подходе, основанном на наземных ориентирах, несколько характерных объектов находятся на плоскости изображения и сверяются по их известному трехмерному местоположению. Используя 2-ые и трехмерные данные, могут быть получены положение и ориентация камеры. Немного примеров для таких алгоритмов - [2], [3]. Как только наземные ориентиры были найдены, нахождение положения и ориентации камеры просто и может быть достигнуто весьма точно. Основная трудность - обнаружение наземных ориентиров и нахождение их правильного соответствия к объектам из известного набора.

В методе интеграции собственного движения, движение камеры относительно себя самой оценивается. Движение эго может быть получено из поля оптического потока, или из приборов, таких как акселерометры и гироскопы. Как только собственное движение было получено, можно интегрировать это движение и найти путь камеры. Один из коэффициентов, которые делают этот подход привлекательным, - то, что никаких наземных ориентиров не нужно находить, в отличие от предыдущего подхода. Несколько алгоритмов оценки собственного движения могут быть найдены в [4], [5], [6], [7].

Недостаток метода интеграции собственного движения происходит из того обстоятельства, что маленькие ошибки суммируются во время процесса интеграции. Следовательно, ошибка предполагаемого положения и ориентации камеры накапливается и происходит уменьшение точности оценки позы с течением времени. Если такой подход используется, было бы желательно уменьшить накопление ошибки, активизируя, время от времени, дополнительный алгоритм, который оценивает позу непосредственно. В [8] предлагается такая навигационная система. В этой работе, также как и в настоящей работе, накопление ошибки исправляется, используя Цифровую Карту Ландшафта (ЦКЛ). ЦКЛ - дискретное представление топографии наблюдаемой местности. Оно содержит высоту над уровнем моря ландшафта для каждого географического местоположения. В [8] кусок местности был восстановлен, используя 'структуру из движения' (СИД) алгоритм, и согласован с ЦКЛ, чтобы получить позу камеры. Использование алгоритма СИД, который не использует информацию, полученную из ЦКЛ, а скорее основывает свою оценку на одном только поле оптического потока, может быть подвержено тому же критическому анализу, что и сам алгоритм СИД [1].

Алгоритм, представленный в этой работе, не требует промежуточной явной реконструкции трехмерного мира. Комбинируя информацию из ЦКМ непосредственно с информацией из изображений, алгоритм хорошо обусловлен и генерирует точные оценки для разумных сценариев и разумных ошибок. В данной работе это требование исследуется, выполняя анализ ошибок на алгоритма, выделенном выше. Определяя соответствующие характеристики этих источников ошибок, получаем выражение для неопределенности позы и движения камеры и затем изучаем влияние различных коэффициентов, используя обширное числовое моделирование.

## 0.2 Определение и описание проблемы.

Проблема может быть кратко описана следующим образом: В любой момент времени $t$, система координат $C(t)$ установлена на камеру таким способом, что $Z$-ось совпадает с оптической осью, и начало координат совпадает с центром проектирования камеры. В этот момент времени камера расположена в некотором географическом местоположении $p(t)$ и имеет данную ориентацию $R(t)$ относительно глобальной координатной системы $W$ ($p(t)$ - трехмерный вектор, $R(t)$ - ортонормальная матрица вращения). $p(t)$ и $R(t)$ определяют преобразование из координатной системы камеры $C(t)$ в глобальную координатную систему $W$, где, если $^C v$ и $^C v$ являются векторами в $C(t)$ и $W$ соответственно, то $^W v = R(t)^C v + p(t)$.

Рассмотрим теперь два последовательных момента времени $t_1$ и $t_2$: преобразование из $C(t_1)$ в $C(t_2)$ дано вектором сдвига $\Delta p(t_1, t_2)$ и матрицей вращения $\Delta R(t_1, t_2)$ таким образом, что $^{C(t_2)} v = \Delta R(t_1, t_2)\, ^{C(t_1)} v + \Delta p(t_1, t_2)$. Для грубой оценки позы камеры в $t_1$ и собственное движение камеры между двумя моментами времени - $p_E(t_1)$, $R_E(t_1)$, $\Delta p_E(t_1, t_2)$ и $\Delta R_E(t_1, t_2)$ используются. (Символ "$E$" для нижнего индекса обозначает, что это оценочная (estimated) величина.)

Также используется поле оптического потока: $\{u_i(t_k)\}$ (i=1 … n, k=1,2). Для $i$'th характерная точка местности, $u_i(t_1) \in \mathbb{R}^2$ и $u_i(t_2) \in \mathbb{R}^2$ представляют её местоположение на первом и втором кадре соответственно.

Используя вышеупомянутые обозначения, цель предложенного алгоритма состоит в том, чтобы оценить истинной позу камеры и её собственное движение: $p(t_1)$, $R(t_1)$, $\Delta p(t_1, t_2)$ и $\Delta R(t_1, t_2)$ используя поле оптического потока $\{u_i(t_k)\}$, ЦКЛ и приблизительные начальные условия: $p_E(t_1)$, $R_E(t_1)$, $\Delta p_E(t_1, t_2)$ и $\Delta R_E(t_1, t_2)$.

## 0.3 Навигационный алгоритм.

Следующий раздел описывает навигационный алгоритм, который оценивает вышеупомянутые параметры. Поза и движение эго камеры получены, используя ЦКЛ и поле оптического потока для двух последовательных кадров. В отличие от метода наземных ориентиров никакие характерные ориентиры не должны быть обнаружены и распознаны. Только соответствие между двумя последовательными изображениями должно быть найдено, чтобы получить поле оптического потока. Как было упомянуто в предыдущем разделе, грубая оценка искомых параметров используется как первое приближение. Однако, так как алгоритм только используют это как начальное приближение и повторно вычисляет позу и движение непосредственно, никакое сложение предыдущих ошибок не будет иметь место, и точность будет сохраняться.

Новый подход основан на следующем наблюдении. Так как ЦКЛ предоставляет информацию о структуре наблюдаемого ландшафта, глубину наблюдаемых характерных точек местности определяет поза камеры. Следовательно, учитывая позу и движение камеры, поле оптического потока может быть однозначно определено. Целью алгоритма является нахождение позы и движения, которые приводят к полю оптического потока наиболее близкому, насколько это возможно, к найденному выше полю оптического потока.

Единичный вектор из поля оптического потока будет использоваться, чтобы определить ограничения на позу камеры и её движение. Пусть $^W G \in \mathbb{R}^3$ является местоположением характерной точки местности в трехмерном пространстве. В два различные моменты времени $t_1$ и $t_2$, эта характерная точки местности проектируется на плоскость изображения камеры в точки $u(t_1)$ и $u(t_2)$. Используя модель дырочной камеры, получаем $u(t_1), u(t_2) \in \mathbb{R}^2$. Позвольте $^{C}q(t_1)$ and $^{C}q(t_2)$ быть гомогенными представлениями этих местоположений. Также можно описать эти вектора как вектора из оптического центров камер к точкам проектирования на плоскости изображений. Используя начальную оценку позы камеры в $t_1$, линия, проходящая через $p_E(t_1)$ и $^{C}q(t_1)$, может быть пересечена с ЦКЛ. Любой алгоритм трассировки лучей может использоваться для этой цели. Местоположение этого пересечения обозначено как $^{W}G_E$. Символ "$E$" нижнего индекса проясняет тот факт, что эта точка местности - предполагаемое местоположение для характерной точки, который вообще будет отличаться от истинного местоположения характерной точки $^{W}G$. Различие между истинным и оцененным местоположением происходит из двух основных источников: ошибка в начальном предположении для позы и ошибка в определении $^{W}G_E$, вызванной дискретизацией ЦКЛ и основными погрешностями. Для разумных ошибок начального положения и ЦКЛ - связанных ошибок, две точки $^{W}G_E$ и $^{W}G$ должны быть достаточно близким, чтобы позволить линеаризацию ЦКЛ вокруг $^{W}G_E$. Обозначая $N$ нормаль плоскости, касательной к DTM в точке $^{W}G_E$, можно написать:

$$N^T(^{W}G - {^{W}G_E}) \approx 0 \qquad (1)$$

Истинная характерная точка местности $^{W}G$ может быть описана, используя истинные параметры позы:

$$^{W}G = R(t_1) \cdot {^{C}q(t_1)} \cdot \lambda \;+\; p(t_1) \qquad (2)$$

Здесь, $\lambda$ обозначает глубину характерной точки (то есть расстояние от точки до плоскости изображения, спроектированное на оптическую ось). Подставляя (2) в (1):

$$N^T(\lambda \cdot R(t_1) \cdot {^{C}q(t_1)} \;+\; p(t_1) \;-\; {^{W}G_E}) = 0 \qquad (3)$$

Из этого выражения глубина истинной характерной точки может быть рассчитана, используя оценочное положение характерной точки:

$$\lambda = \frac{N^{T\,W}G_E - N^T p(t_1)}{N^T R(t_1) {^{C}q(t_1)}} \qquad (4)$$

Подстановкой (4) обратно в (2) получаем:

$$^{W}G = R(t_1) {^{C}q(t_1)} \cdot \left( \frac{N^{T\,W}G_E - N^T p(t_1)}{N^T R(t_1) {^{C}q(t_1)}} \right) + p(t_1) \qquad (5)$$

Чтобы упростить обозначения, $R(t_i)$ будет заменен $R_i$ и аналогично для $p(t_i)$ и $q(t_i)$ $i = 1, 2$. $\Delta R(t_1, t_2)$ и $\Delta p(t_1, t_2)$ будут заменены на $R_{12}$ и $p_{12}$ соответственно. Верхний индекс, описывающий систему координат, в которой дан вектор, будет также опущен, за исключением случаев, где требуется особое внимание к описываемым системам координат. Обычно, $p_{12}$ и $q$'s находится в системе координат камеры, в то время как остальная часть векторов дана в глобальной системе координат. Используя упрощенные обозначения, (5) может быть переписано как:

$$G \;=\; \frac{R_1 q_1 N^T}{N^T R_1 q_1} G_E \;-\; \frac{R_1 q_1 N^T}{N^T R_1 q_1} p_1 \;+\; p_1 \qquad (6)$$

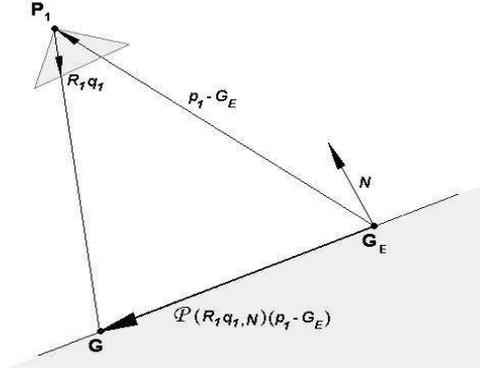

Рис. 1: Геометрическое описание выражения (9) используя проектирующий оператор (7)

Для того, чтобы получить более простое выражение определим следующий проектирующий оператор:

$$\mathcal{P}(u, s) \doteq \left(\mathrm{I} - \frac{us^T}{s^T u}\right) \tag{7}$$

Этот оператор проектирует вектор на нормаль подпространства к $s$, вдоль направления $u$. Как иллюстрация, это просто проверить, что $s^T \cdot \mathcal{P}(u, s)v \equiv 0$ и $\mathcal{P}(u, s)u \equiv 0$. Добавляя и вычитая $G_E$ к (6), после переупорядочения:

$$G = G_E + \left[\mathrm{I} - \frac{R_1 q_1 N^T}{N^T R_1 q_1}\right] p_1 - \left[\mathrm{I} - \frac{R_1 q_1 N^T}{N^T R_1 q_1}\right] G_E \tag{8}$$

Используя проектирующий оператор, (8) становится:

$$G = G_E + \mathcal{P}(R_1 q_1, N)\left(p_1 - G_E\right) \tag{9}$$

У вышеописанного выражения есть ясная геометрическая интерпретация (см. Рис.1). Вектор из $G_E$ к $p_1$ проектируется на касательную плоскость. Проектирование идет вдоль направления $R_1 q_1$, которое является направлением луча из оптического центра камеры $(p_1)$, проходящего через соответствующую точку изображения.

Наш следующий шаг будет перенос $G$ из глобальной системы координат - $W$ в систему координат первой камеры $C_1$ и затем к системе координат второй камеры $C_2$. Так как $p_1$ and $R_1$ описывают преобразование из $C_1$ в $W$, мы будем использовать обратное преобразование:

$$^{c_2}G = p_{12} + R_{12}\left(R_1^T\left(G - p_1\right)\right) \tag{10}$$

Подстановка (9) в (10) дает:

$$^{c_2}G = p_{12} + R_{12}\mathcal{L}\left(G_E - p_1\right) \tag{11}$$

$\mathcal{L}$ в написанном выше выражении представляет:

$$\mathcal{L} = \frac{q_1 N^T}{N^T R_1 q_1} \tag{12}$$

Можно интерпритировать $\mathcal{L}$ как оператор обратный к $\mathcal{P}$: он проектирует векторы на луч, продолжающий $R_1 q_1$ вдоль плоскости, ортогональной к $N$.

$q_2$ - проекция истинной характерной точки местности $G$. Таким образом, векторы $q_2$ и $^{c_2}G$ должны совпасть. Это наблюдение может быть выражено математически, проектируя $^{c_2}G$ на продолжение луча $q_2$:

$$^{c_2}G = \frac{q_2}{|q_2|} \cdot \left( \frac{q_2^T}{|q_2|} \cdot {}^{c_2}G \right) \qquad (13)$$

В выражении (13), $q_2^T/|q_2| \cdot {}^{c_2}G$ является величиной $^{c_2}G$'s прекции на $q_2$. Преобразуя (13) и используя проектирующий оператор, мы получаем:

$$\left[ \mathrm{I} - \frac{q_2 \cdot q_2^T}{q_2^T \cdot q_2} \right] \cdot {}^{c_2}G \quad = \quad \mathcal{P}(q_2, q_2) \cdot {}^{c_2}G \quad = \quad 0 \qquad (14)$$

$^{c_2}G$ является проекцией на ортогональную компонент $q_2$. Так как $^{c_2}G$ и $q_2$ должны совпадать, это проектирование должно давать нулевой вектор. Подстановка (11) в (14) приводит к нашему окончательному ограничению:

$$\mathcal{P}(q_2, q_2) \left[ p_{12} + R_{12} \mathcal{L} \left( G_E - p_1 \right) \right] = 0 \qquad (15)$$

Это ограничение включает позицию, ориентацию и собствнное движение, определяемое на основе двух кадров камеры. Хотя оно включают трехмерные вектора, ясно, что его ранг не может превысить двойку из-за использования $\mathcal{P}$, который проектирует $\mathbb{R}^3$ на двумерное подпространство.

Такое ограничение может быть установлено для каждого вектора в поле оптического потока, пока не будет получена несингулярная система. Так как двенадцать параметров должны быть оценены (шесть для позы и шесть для собственного движения), по крайней мере, шесть векторов оптического потока требуются для решения системы. Но это - правильное заключение для нелинейной проблемы. Если мы используем метод итераций Гаусса-Ньютона, то делаем линеаризацию нашей проблемы около приближенного решения. Найденная матрица будет всегда сингулярной для шести точек (с нулевым детерминантом), как численное моделирование демонстрирует. Таким образом, необходимо использовать, по крайней мере, семь точек, чтобы получить несингулярное линейное приближение. Обычно, чем больше векторов будет использоваться, чтобы определить переопределенную систему, тем более устойчиво решение. Внимание должно быть привлечено к факту, что было получено нелинейное ограничение. Таким образом, итерационная схема будет использоваться, чтобы решить эту систему. Устойчивый алгоритм, который использует итерации Гаусса-Ньютона и М-оценщик описан в [9]. Мы начинаем использовать метод Levenberg-Marquardt, если метод Гаусса-Ньютона после нескольких итераций прекратил сходиться. Эти два алгоритма реализованы в функции lsqnonlin() пакета Matlab. Применимость, точность и надежность алгоритма были проверены через численное моделирование и лабораторные эксперименты.

Более удобно использовать более устойчивое для итераций решение, эквивалентное уравнению (15):

$$\mathcal{P}(q_2, q_2) \left[ p_{12} + R_{12} \mathcal{L}_i \left( G_{E_i} - p_1 \right) \right] / |{}^{c_2}G| = 0 \qquad (16)$$

Использование этой нормализованной формы уравнений предотвращает получение неверного тривиального решения, когда две позиции находятся в единой точке на местности.

## 0.4 Навигационный алгоритм, основанный на компьютерном зрении, для коррекции инерциальной навигации с помощью фильтра Калмана.

Основанные на компьютерном зрении навигационные алгоритмы были главной исследовательской проблемой в течение прошлых десятилетий. Алгоритм, используемый в этой статье, основан на геометрии многих изображений и карте местности. С помощью этого метода мы получаем позицию и ориентацию наблюдающей камеры. С другой стороны мы получаем те же самые данные из инерциальных навигационных методов. Чтобы скорректировать эти два результата, используется фильтр Калмана. Мы используем в этой статье расширенный фильтр Калмана для нелинейных уравнений [12].

Для инерциальных навигационных вычислений использовался Инерциальный Навигационный Системный Пакет для Matlab [13].

Вход фильтра Калмана состоит из двух частей. Первый - переменные $X$ для уравнений движения. В нашем случае это - инерциальные навигационные уравнения. Вектор $X$ состоит из пятнадцати компонентов: $[\delta x \delta y \delta z \delta V_x \delta V_y \delta V_z \delta \phi \delta \theta \delta \psi a_x a_y a_z b_x b_y b_z]$. Координаты $\delta x \delta y \delta z$ определены как разница между реальной позицией камеры и позицией, полученный из инерциального навигационного вычисления. Переменные $\delta V_x \delta V_y \delta V_z$ определены как разница между реальной скоростью камеры и скоростью, полученной из инерциального навигационного вычисления. Переменная $\delta \phi \, \delta \theta \, \delta \psi$ определена как углы Эйлера матрицы $D_r * D_c^T$. Здесь $D_r$ матрица определена реальными углами Эйлера камеры относительно локальной системы координат (L-frame). С другой стороны $D_c$ матрица определена углами Эйлера камеры относительно локальной системы координат (L-frame), полученными из инерциального навигационного вычисления. Необходимо обратить внимание, что найденные углы Эйлера $\delta \phi \, \delta \theta \, \delta \psi$ НЕ эквивалентны разнице между реальными углами Эйлера и углами Эйлера, полученными из инерциального навигационного вычисления. Однако для маленьких значений $\delta \phi \, \delta \theta \, \delta \psi$ попрвки к этим углам могут быть добавлены линейно и таким образом эти углы могут использоваться в фильтре Калмана в случае маленьких ошибок. Такой выбор углов сделан, поскольку формулы, описывающие их эволюцию, намного более просты, чем формулы, описывающие эволюцию разницы между углами Эйлера. Переменные $a_x a_y a_z$ определены вектором смещения ускорения в инерциальных навигационных измерениях. Переменные $b_x b_y b_z$ определены вектором гироскопического смещения в инерциальных навигационных измерениях.

Второй вход фильтра Калмана - $Z$-результат измерений, основанных на навигационных алгоритмах компьютерного зрения. Вектор $Z$ состоит из шести компонентов $[\delta x_m \; \delta y_m \; \delta z_m \; \delta \phi_m \; \delta \theta_m \; \delta \psi_m]$. Координаты $\delta x_m \; \delta y_m \; \delta z_m$ являются разницей между позицией камеры, измеренной на основе навигационного алгоритма компьютерного зрения, и позицией, полученный из инерциального навигационного вычисления. Переменная $\delta \theta_m \; \delta \psi_m$ определена как углы Эйлера матрицы $D_m * D_c^T$. Здесь $D_m$ матрица определена углами Эйлера камеры относительно относительно локальной системы координат (L-frame), измеренных на основе навигационного алгоритма компьютерного зрения. С другой стороны $D_c$ матрица определена углами Эйлера камеры относительно локальной системы координат (L-frame), полученными из инерциального навигационное вычисления. Пусть переменная $k$ определяет число шагов для дискретизации времени, используемых в фильтре Калмана.

Мы полагаем, что ошибки между значениями, плученными из инерциаль-ного навигационного вычисления, и реальными значениями линейно зависят от шума. Соответствующая ковариационная матрица шума обозначена $Q_k$. Диагональные элементы $Q_k$ соответствуют скорости, определяются шумом ускорения и пропорциональны $dt^2$: $Q_V \sim dt^2$, где $dt$ интервал времени между $t_k$ и $t_{k-1}$: $dt = t_k - t_{k-1}$. Диагональные элементы $Q_k$ соответствуют углам Эйлера, определяются гироскопическим шумом и пропорциональны $dt$: $Q_A \sim dt$.

Мы предполагаем, что также ошибки между значениями, полученными на основе навигационного алгоритма компьютерного зрения, и реальными значениями линейно зависят от шума. Соответствующая ковариационная матрица шума обозначена $R_k$. Анализ ошибок, дающий эту матрицу, описан в [14].

Уравнения фильтра Калмана описывают как эволюцию оценок состояния $X_k$, описанных выше, так и эволюцию ковариационной матрицы $P_k$ для переменных $X_k$.

Чтобы написать уравнения фильтра Калмана, мы должны определить еще две 15х15 матрицы: $H_k$ и $A_k$. Матрица $H_k$ является якобианом измерения, описывающая связь между предсказанным измерением $H_k * X_k$ и фактическим измерением $Z_k$, определенным выше. Диагональные элементы $H_k(1,1)$, $H_k(2,2)$, $H_k(3,3)$, описывающие координату и элементы $H_k(4,7)$, $H_k(5,8)$, $H_k(6,9)$, описывающие углы, равны единице. Остальная часть элементов равна нулю.

$A_k$ матрица Якоби, описывающая эволюцию вектора $X_k$. Точное выражение для этой матрицы является очень сложным, поэтому мы используем приближенную формулу для $A_k$, пренебрегая эффектами кориолиса, вращением Земли и так далее. Позвольте $\phi\ \theta\ \psi$ быть углами Эйлера в L-frame, $dV$ является вектором deltaV, полученным из инерциальных навигационных измерений, $f_{vec}$ вектор ускорения в L-frame, $DCM_\text{b-to-l}$ - матрица направляющих косинусов (для перевода из координатной системы летательного аппарата в L-frame).

Формулы определяющие $A_k$, следующие:

$$\Psi_{DCM} = \begin{pmatrix} \cos(\psi) & \sin(\psi) & 0 \\ -\sin(\psi) & \cos(\psi) & 0 \\ 0 & 0 & 1 \end{pmatrix} \tag{17}$$

$$\Theta_{DCM} = \begin{pmatrix} \cos(\theta) & 0 & -\sin(\theta) \\ 0 & 1 & 0 \\ \sin(\theta) & 0 & \cos(\theta) \end{pmatrix} \tag{18}$$

$$\Phi_{DCM} = \begin{pmatrix} 1 & 0 & 0 \\ 0 & \cos(\phi) & \sin(\phi) \\ 0 & -\sin(\phi) & \cos(\phi) \end{pmatrix} \tag{19}$$

$$DCM_\text{b-to-l} = \Phi_{DCM}\Theta_{DCM}\Psi_{DCM} \tag{20}$$

$$f_{vec} = DCM_\text{b-to-l}\frac{dV}{dt} \tag{21}$$

$$Phi(1:3, 4:6) = \begin{pmatrix} 1 & 0 & 0 \\ 0 & 1 & 0 \\ 0 & 0 & 1 \end{pmatrix} \tag{22}$$

$$Phi(4:6, 7:9) = \begin{pmatrix} 0 & -f_{vec}(3) & f_{vec}(2) \\ f_{vec}(3) & 0 & -f_{vec}(1) \\ -f_{vec}(2) & f_{vec}(1) & 0 \end{pmatrix} \tag{23}$$

$$Phi(7:9, 10:12) = -DCM_{\text{b-to-l}} \qquad (24)$$

$$Phi(4:6, 13:15) = -DCM_{\text{b-to-l}} \qquad (25)$$

Остальные элементы матрицы Phi равны нулю.

$$A_k = I + Phi\, dt \qquad (26)$$

Уравнения фильтра Калмана для временной эволюции следующие

$$X_k^- = [0\,0\,0\,0\,0\,0\,0\,0\,0\, a_{xk-1}\, a_{yk-1}\, a_{zk-1}\, b_{xk-1}\, b_{yk-1}\, b_{zk-1}] \qquad (27)$$

$$P_k^- = A_k P_{k-1} A_k^T + Q_{k-1} \qquad (28)$$

Уравнения фильтра Калмана проектируют состояние и ковариационную матрицу с предыдущего временного шага $k-1$ на текущий временной шаг $k$.

Уравнения фильтра Калмана для измерения следующие:

$$K_k = P_k^- H_k^T (H_k P_k^- H_k^T + R_k)^{-1} \qquad (29)$$

$$X_k = X_k^- + K_k(Z_k - H_k X_k^-) \qquad (30)$$

$$P_k = (I - K_k H_k) P_k^- (I - K_k H_k)^T + K_k R_k K_k^T \qquad (31)$$

Уравнения фильтра Калмана для измерения исправляют состояние и ковариационной матрицу в соответствии с измерением $Z_k$.

Найденный вектор $X_k$ используется, чтобы обновить координаты, скорости, углы Эйлера, смещение ускореня и гироскопическое смещение для инерциальных навигационных вычислений на следующем шаге.

Численные расчеты были реализованны, чтобы исследовать эффективность фильтра Калмана и чтобы объединить эти два навигационных алгоритма. На Рис. 2 мы можем видеть, что для откорректированного пути ошибка координаты, полученные на основе двух навигационных методов с фильтрацией Калмана, намного меньше чем инерциальная навигационная ошибка координаты, полученная без фильтра Кальмана. Улучшенные результаты с помощью фильтра Калмана были получены также для скорости, несмотря на то, что эта скорость не измеряется напрямую навигационным алгоритмом, использующим компьютерное зрение Рис. 3.

## 0.5 Выводы

Алгоритм для оценки позы и движения, использующий соответствующие точки в изображениях и ЦКЛ был представлен с использованием фильтра Калмана. ЦКЛ служит как глобальная справочная информация, и ее данные используются для того, чтобы восстановить абсолютную позицию и ориентацию камеры. В численных расчетах оценка для позиции и скорости находятся достаточно точно, чтобы предотвратить накопленные ошибки и предотвратить дрейф траектории.

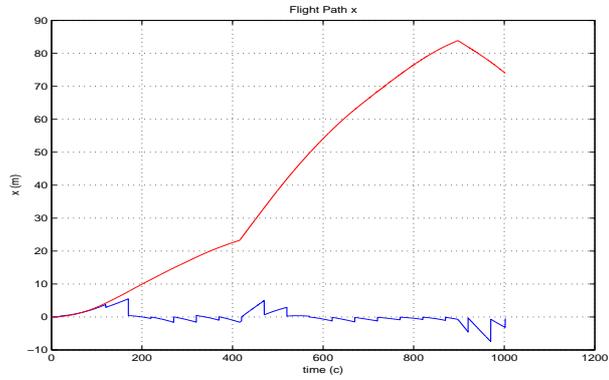

(a)

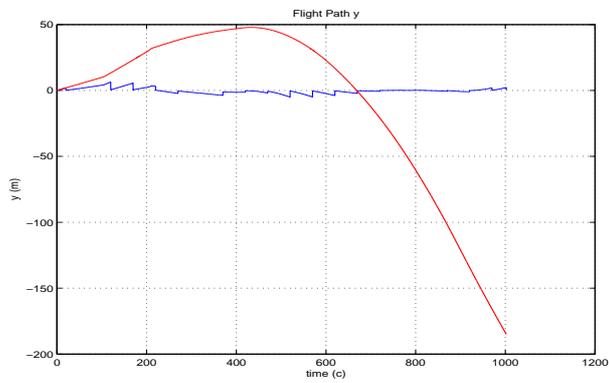

(b)

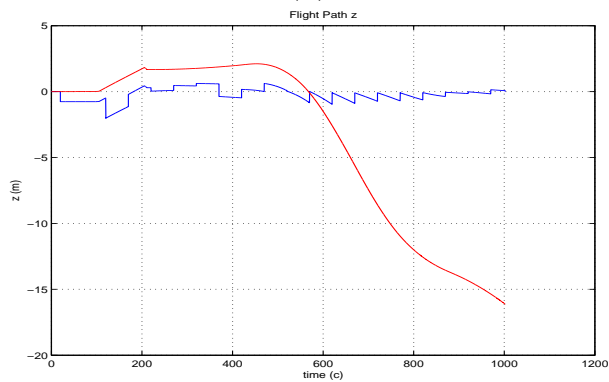

(c)

Рис. 2: Ошибки позиции ((a) для координаты x (b) для координаты y (c) для координаты z). Ошибки инерциального дрейфа отмечены красной линией, и ошибки, исправленные видео-навигацией, отмечены синей линией.

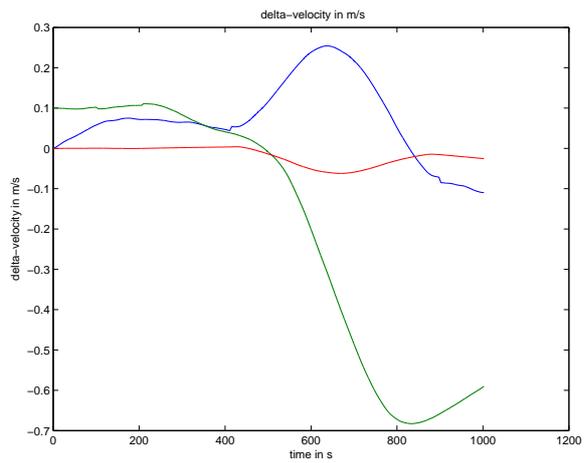

(a)

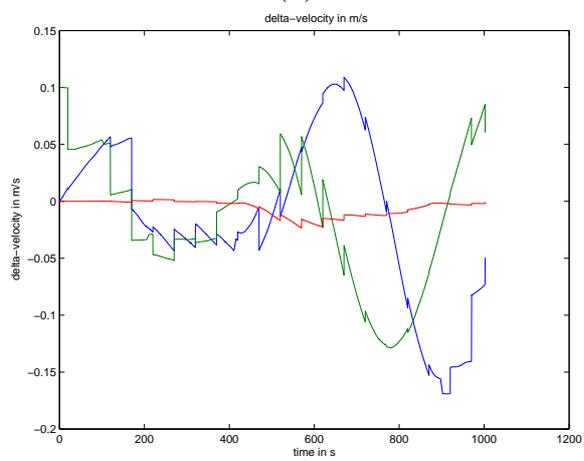

(b)

Рис. 3: (a) ошибки скорости для инерциального дрейфа (x y z компоненты), и (b) ошибки скорости, исправленные видео-навигацией (x y z компоненты).

# Литература